\pdfoutput=1

\documentclass[11pt]{article}
\usepackage[table,xcdraw]{xcolor}
\usepackage[]{EMNLP2023}

\usepackage{times}
\usepackage{latexsym}

\usepackage[T1]{fontenc}

\usepackage[utf8]{inputenc}

\usepackage{microtype}

\usepackage{inconsolata}
\usepackage{graphicx}
\usepackage{booktabs}
\usepackage{multirow}

\usepackage{color}
\definecolor{wildstrawberry}{rgb}{1.0, 0.26, 0.64}

\usepackage{times}
\usepackage{latexsym}
\usepackage{xspace}
\usepackage{booktabs}
\usepackage{graphicx}
\usepackage{multirow}
\usepackage{tikz}
\usetikzlibrary{plotmarks}
\usepackage{fontawesome5}
\usepackage[tight,footnotesize]{subfigure}

\usepackage{algorithm}
\usepackage{algorithmic}
\usepackage{soul}
\usepackage{enumitem}
\usepackage{todonotes}

\newcommand{\RNum}[1]{\uppercase\expandafter{\romannumeral #1\relax}}
\newcommand{\SideNote}[2]{\todo[color=#1,size=\footnotesize]{#2}}
\newcommand{\maxside}[1]{\SideNote{red!40}{#1 --Max}}
\newcommand{\maxma}[1]{\textcolor{red}{[#1]}}

\newcommand{\mkclean}{
\renewcommand{\maxma}[1]{}
\renewcommand{\maxside}[1]{}
}
\mkclean

\title{Evaluating Large Language Models on Controlled Generation Tasks}

\author{
Jiao Sun$^{1}$\thanks{\xspace\xspace The first four authors contribute equally.} \,\, \textbf{Yufei Tian}$^{2*}$ \, \textbf{Wangchunshu Zhou}$^{3*}$ \, \textbf{Nan Xu}$^{1*}$
\\ 
\textbf{Qian Hu}$^4$ \, \textbf{Rahul Gupta}$^4$ \, \textbf{John Wieting}$^5$ \, \textbf{Nanyun Peng}$^2$ \, \textbf{Xuezhe Ma}$^1$ \
\\
$^1$University of Southern California \,
$^2$University of California, Los Angeles \\
$^3$ ETH Zurich \,
$^4$ Amazon \,
$^5$ Google DeepMind
\\
\texttt{\{jiaosun,nanx,xuezhema\}@usc.edu} \, \texttt{\{yufeit,violetpeng\}@cs.ucla.edu}
\\
\texttt{wangchunshu.zhou@inf.ethz.ch} \, \texttt{\{huqia, gupra\}@amazon.com} \\\texttt{jwieting@google.com}
}

\begin{document}
\maketitle
\begin{abstract}
While recent studies have looked into the abilities of large language models in various benchmark tasks, few studies have looked into the controllability of large language models on generation tasks. We present a systematic and extensive analysis of the controllability of large language models on ten benchmarks, including a new simple yet challenging numerical planning benchmark with different granularities. After comparing large language models against state-of-the-start finetuned smaller models, we present a spectrum showing when large language models fall behind, are comparable, or exceed the ability of smaller models. We conclude that \textbf{large language models struggle at meeting fine-grained hard constraints}.
\end{abstract}

\section{Introduction}
Text generation models should generate texts that meet controllable constraints as humans wish~\cite{Zhang2022ASO}.   
For example, one can avoid the blandness caused by repetitive patterns by controlling the syntax of generated sentences~\cite{iyyer-etal-2018-adversarial,qian-etal-2019-exploring}. In a customized dialogue system, one should be able to control the persona of the utterance~\cite{smith2020controlling}. 
Previous works either finetune generation models such as BART~\cite{Lewis2019BARTDS} on specific tasks for better controllability (e.g., controlled paraphrase generation~\cite{sun-etal-2021-aesop}) or design constrained decoding strategies (e.g., look-back decoding strategy by \citet{xu2023lookback}) for controlled generation.

Large Language Models (LLMs) have recently shown great potential in various generation tasks. 
For example, ~\citet{chatgptformt} shows that \texttt{ChatGPT} with GPT-4 as an engine achieves commercial-level machine translation quality. 
~\citet{chatgpt-benchmark} find that annotators prefer summaries generated from \texttt{ChatGPT} over state-of-the-art summarization models. 
However, few works investigate the controllability of large language models.
Towards this end, we aim to study and understand the controllability of large language models to answer the question: \emph{Are large language models better than finetuned smaller models at controllability on generation tasks?}. 

\begin{figure}
    \centering
    \includegraphics[width=\linewidth]{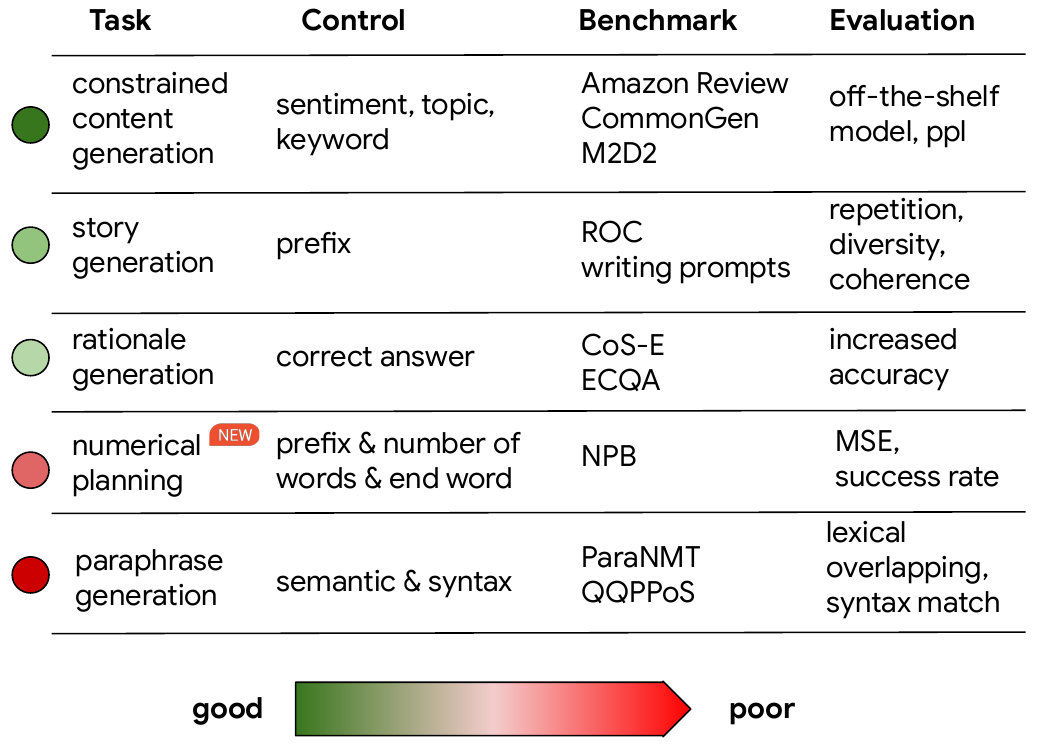}
    \caption{We test large language models on five controlled generation tasks with various control factors using automatic evaluation methods. We show a spectrum of abilities of large language models on such tasks and conclude that large language models struggle at fine-grained hard constraints such as numerical planning.}
    \label{fig:illustration}
\end{figure}

The main contribution of this work is to conduct a comprehensive analysis of LLM's controllability on five tasks and ten generation benchmarks, including controlled story generation, controlled free-form generation with sentiment and topics, controlled paraphrase generation, and controlled rationale generation as in Figure~\ref{fig:illustration}. 
We further design a new simple yet challenging benchmark named Numerical Planning Benchmark (NPB), where the task is to satisfy numerical constraints from four granularities (word-, syllable-, sentence- and paragraph-level) and under different content controls (e.g., prefix and ending).
For evaluation, we use automatic metrics, which are imperfect yet convenient and reproducible.\footnote{\url{https://github.com/sunjiao123sun/llm-controlgen}}

After an in-depth examination, we categorize LLM's controllability on a spectrum: from lagging behind and being on par with to surpassing smaller finetuned models. Our findings indicate that large language models have difficulties adhering to specific hard constraints, such as numerical planning. 

We first introduce the numerical planning task and the associated evaluation as this is a new, intuitively simple, yet challenging task (\S\ref{sec:sentence_planning}). For the rest, we rank them by the task difficulty indicated in Figure~\ref{fig:illustration} from easy to hard: constrained content generation   (\S\ref{sec:content_generation}), story generation (\S\ref{sec:story_generation}), rationale generation (\S\ref{sec:rationale-generation}) and paraphrase generation (\S\ref{sec:paraphrase_generation}).
 
\vspace{-1mm}
\section{Numerical Planning}
\label{sec:sentence_planning}
\begin{quote}
    \emph{Can LLMs count from two to ten?}\vspace{-2mm}
\end{quote}
\paragraph{Task Description.}

We introduce the Numerical Planning Benchmark (NPB) as an intuitive task that tests the \textit{basic numerical planning ability} of LLMs. The high-level task descriptions can be found in Table \ref{tab:NPB_illustration}. We are inspired by real-world scenarios such as creative writing. For example, writers may wish to generate sentences or poems with a specific structure, such as a fixed number of words or syllables in each line, aiming to adhere to particular forms (\textit{e.g.,} sonnets, where each line contains exactly 10 or 11 syllables \cite{tian-peng-2022-zero}). Meanwhile, humans may also want full control over the start and end of each line for rhetorical purposes such as alliteration and rhyming. Inductively, we formulate our numerical planning benchmark from four different granularities: generating a piece of text that contains a predefined number of \textit{words, syllables, sentences, or paragraphs} given a plausible pair of prefix (start) and suffix (ending) as constraints. The prefix is given to LLMs such that they are only queried to generate the continuations.

\begin{table}[]
\small
\centering
\begin{tabular}{@{}l|l@{}}
\toprule
\textbf{Granularity} & \textbf{Task Illustration}                                                                                                                        \\ \midrule
\multirow{3}{*}{Word/Syllable}  &  \begin{tabular}[c]{@{}l@{}}Generate a sentence using exactly 5 \\words/syllables.\end{tabular}                                                                                                          \\ \cmidrule(l){2-2} 
                       & \begin{tabular}[c]{@{}l@{}}Complete sentence ``This is a story''\\ using exactly 5 words/syllables.\end{tabular}                                     \\ \cmidrule(l){2-2} 
                       & \begin{tabular}[c]{@{}l@{}}Complete sentence ``This is a story''\\ using exactly 5 words/syllables, \\ including the last word as ``town''.\end{tabular} \\ \midrule
Sentence               & Generate a paragraph with 5 sentences, ...                                                                                                                             \\ \midrule
Paragraph              & Generate an article with 5 paragraphs, ...                                                                                                                            \\ \bottomrule
\end{tabular}
\caption{Task illustration for the Numerical Planning Benchmark. We test LLMs' numerical planning ability under various constraints (word counting and end word) and granularities (word, syllable, sentence, and paragraph). Due to space limitations, we only show the full constraints under the word granularity here.}
\label{tab:NPB_illustration}
\end{table}

\paragraph{Evaluation Metrics.} We use success rate (SR) and mean squared error (MSE) as automatic evaluation metrics. As our control is two-fold, we separately calculate the success rates of 1) generating the continuation with the correct counts and 2) generating the continuation with the proper ending. We also calculate the MSE between our input numbers and output numbers.

\begin{figure*}
    \centering
    \includegraphics[width=\linewidth]{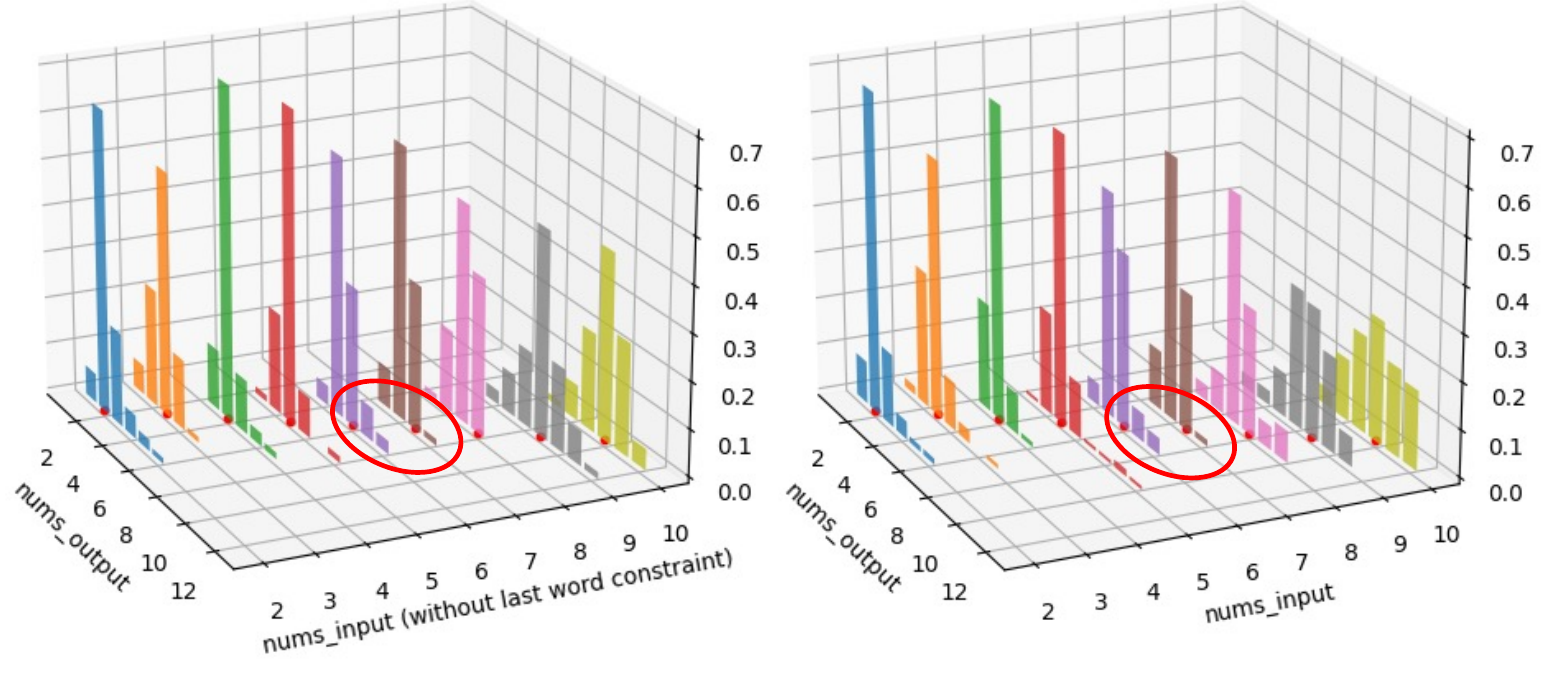}
    \caption{Histogram visualization in the distribution (frequency, z-axis) of input numbers (x-axis) and output numbers (y-axis) for word count planning. Left: querying \texttt{ChatGPT} to generate a continuation of a given prefix with $N$ words. Right: querying \texttt{ChatGPT} to generate a continuation with $N$ words of a given prefix that ends with a given word. Small red dots \textcolor{red}{\textbullet} mark those bars where output numbers equal input numbers. These bars represent the fine-grained success rates. For either case, there is a significant drop when the input number reaches six.}
    \label{fig:3d_plot}
\end{figure*}

\paragraph{Evaluate with LLMs.} We evaluate \texttt{ChatGPT} and \texttt{Alpaca-7b} on our NPB benchmark in zero-shot and few-shot settings. Each request used to query the LLMs corresponds to a real case in the datasets of Romance Books and Reddit Short Stories.\footnote{huggingface.co/datasets/AlekseyKorshuk/romance-books, www.kaggle.com/datasets/trevordu/reddit-short-stories} For word-level planning tasks (word and syllable count), we randomly select sentences from the above datasets. Then, we select the last word in each sentence as the suffix. Depending on how many additional words we query the LLMs to generate, we select the first few words in each sentence as the prefix (if we simply ask LLMs to generate freely without a prefix, the outputs lack diversity). Our prompt is written as \textit{Complete a sentence that starts with \{prefix\} using exactly \{N\} additional words (including the last word \{last word\}). The sentence must end with the word \{last word\}. Sentence: \{prefix\}}, and LLMs will continue. In the few-shot setting, we provide the task description and three examples. For each example, we also provide explanations to help LLMs better understand our task. For example,
\begingroup
\addtolength\leftmargini{-0.2in}
\begin{quote}
    \textit{\#\#Prefix: This is a story about a young girl's \\
\#\#Last word: town \\
\#\#N: 5 \\
\#\#Output: This is a story about a young girl's redemption in a small town. \\
\#\#Explanation: We generated ``redemption in a small town''. It contains exactly 5 words and ends with the last word `town'.}
\end{quote}
\endgroup

\noindent We query the LLMs to generate outputs from $N=2$ to $N=10$ words. Each number $N$ has 100 evaluation samples. For paragraph-level tasks, the prefix and suffix are the first and last sentences in the corresponding paragraphs. For all experiments, our decoding strategy is top $p$ ($p=0.95$) sampling with temperature $T=0.3$ unless otherwise specified. 

\begin{table}[t]
\small\centering
\begin{tabular}{@{}lllll@{}}
\toprule
\textbf{Model} & \textbf{\begin{tabular}[c]{@{}l@{}}SR -\\ count\end{tabular}} & \textbf{\begin{tabular}[c]{@{}l@{}}SR -\\ last word\end{tabular}} & \textbf{\begin{tabular}[c]{@{}l@{}}SR -\\ both\end{tabular}} & \textbf{\begin{tabular}[c]{@{}l@{}}MSE -\\ count\end{tabular}} \\ \midrule
GPT-2 (fine-tuned)  & 0.64                                                          & 0.86                                                              & 0.60                                                         & 1.62 \\
\midrule
Alpaca-7b$_{\texttt{zs}}$      & 0.17                                                          & 0.31                                                              & 0.09                                                         & 9.19                                                           \\
Alpaca-7b$_{\texttt{ICL}}$  & 0.14                                                          & 0.34                                                              & 0.07                                                         & 9.76                                                           \\ 
Vicuna$_{\texttt{zs}}$      & 0.08   & 0.09    & 0.03    &   27.68             \\
Vicuna$_{\texttt{ICL}}$     & 0.13   & 0.30    & 0.04    &   13.43             \\
Falcon$_{\texttt{zs}}$      & 0.13   & 0.42    & 0.08    &   11.60             \\
Falcon-7b$_{\texttt{ICL}}$  & 0.11   & 0.34    & 0.03    &   13.72             \\
\midrule
\texttt{ChatGPT}        & \textbf{0.41}                                                 & 0.74                                                              & \textbf{0.36}                                                & \textbf{3.64}                                                  \\
\texttt{ChatGPT}$_{\texttt{ICL}}$   & 0.37                                                          & \textbf{0.78}                                                     & 0.34                                                         & 4.95                                                           \\
\bottomrule
\end{tabular}
\caption{Success rates for the word count planning task. Surprisingly, few-shot in-context learning (ICL) underperforms zero-shot (zs) on numerical planning.}
\label{tab:word_count_planning}
\vspace{-4mm}
\end{table}

\paragraph{Result.}
We report the model performance of LLMs and a fine-tuned GPT-2-large model on the task of word count planning in Table \ref{tab:word_count_planning}. Due to space limitations, we compile the results of the remaining tasks in Appendix \ref{appendix:addtional_SPB}. First, it is clear LLMs are poor at numerical planning, although it is an extremely simple task for humans. Given its extremely poor performance, we consider \texttt{Alpaca} incapable of doing numerical planning. Secondly, LLMs learn to incorporate literal constraints, such as the last word, via few-shot in-context learning. Interestingly, \textbf{\textit{few-shot in-context learning deteriorates the performance of numerical planning}}. Upon further inspection, we find that LLMs try to mimic the style or features (such as length) in the in-context examples and are, 
therefore, more likely to generate outputs with the wrong word counts once the input number $N$ cannot be found in the examples. Our results resonate with \citet{yin2023did,kung2023models,sinha-etal-2023-language} that LMs do not truly understand task definitions via in-context learning.

Figure \ref{fig:3d_plot} is a fine-grained visualization of the input and output numbers distribution by zero-shot \texttt{ChatGPT}. Specifically, we compare LLMs' numerical planning abilities with (e.g., \textit{complete sentence with ``redemption in a small town'' using exactly 5 words, including the last word as ``happy''}) and without additional suffix constraint (e.g., \textit{complete sentence with ``redemption in a small town'' using exactly 5 words}). LLMs can generate more freely without suffix constraints to meet the numerical constraint. However, \texttt{ChatGPT} doesn't always translate to a higher success rate. We find out that only when $N$ is small (i.e., 2 and 3), \texttt{ChatGPT} achieves a higher success rate if explicitly told the last word of the target sentence. 

Finally, we would like to point out a few behaviors. First, although the general trend is that LLMs' numerical planning ability drops as $N$ increases, $N=3$ is a clear exception (performs worse) among various experiments we repeated. Second, by checking the failure cases, we find that \textit{\texttt{ChatGPT} always generates shorter continuations} than required. Moreover, we see a sudden drop in model performances (from above $\sim$0.6 to $\sim$0.4) when the input number $N$ increases from 5 to 6. We encourage future research to investigate these behaviors.

\section{Content-Controlled Generation}
\label{sec:content_generation}

\paragraph{Task Description.}
We consider three types of content constraints: topic, sentiment, and keyword. The detailed task definitions and dataset can be found in Appendix \ref{appendix:addtional_content_control}.

\paragraph{Evaluation Metrics.}
We use the success rate as the evaluation metric to measure how well LLMs can follow the content constraints. Specifically, we use GPT-3.5~\citep{ouyang2022training} based topic/sentiment classifiers with in-context learning using five examples per category to evaluate whether the generated texts belong to the specified topic or sentiment class. We consider an LLM to succeed in one example if the predicted class of the generated text is identical to the input constraint. For a keyword-constrained generation, we use the keyword coverage metric that measures the percentage of input keywords included in generated texts.

\paragraph{Evaluate with LLMs.}
For the content constrained generation with LLMs, we follow~\citet{zhou2023controlled} and use natural language instructions to prompt LLMs. Specifically, we use a prompt template of \textit{``Write a sentence about \{topic name\}''} for topic-constrained generation, \textit{``Write an Amazon review with \{level number\} star about a random thing. The number of stars ranges from one to five. One star is the most negative, and five stars are the most positive''} for sentiment constraints, and \textit{``Write a sentence using the following keywords: \{keywords\}''} for keyword constraints.

In addition to zero-shot evaluation, we also evaluate LLMs in the in-context learning setting by appending the following demonstration template: \textit{``Below are some examples for the task: Input: \{input 1\}, Output: \{output 1\}; Input: \{input 2\}, Output: \{output 2\} ... ''}. We use 5 in-context examples per class following the practice in~\citet{zhou2023controlled}.

We compare various LLMs including \texttt{ChatGPT}, LLaMA, Alpaca, Vicuna, and Falcon in our experiments. We also report the results of Diffusion-LM~\citep{li2022diffusionlm} based on BERT-large~\citep{devlin-etal-2019-bert} and task-specific classifiers as a competitive non-LLM baseline 

\begin{table}[]
\centering
\small
\begin{tabular}{@{}lccc@{}}
\toprule
\textbf{Model} & \textbf{Topic} & \textbf{Sentiment} & \textbf{Keyword} \\ \midrule
Diffusion-LM & 68.9 & 83.7 & 93.2 \\
GPT-2 (1.5B, fine-tuned) & 63.4 & 76.5 & 88.9 \\
T5 (3B, fine-tuned) & 67.3 & 83.9 & 94.8 \\
\midrule
$\texttt{LLaMA-7B}_{\texttt{zs}}$ & 45.3 & 58.4 & 83.5 \\
$\texttt{LLaMA-7B}_{\texttt{ICL}}$ & 63.5 & 85.1 & 93.0  \\
$\texttt{Alpaca-7B}_{\texttt{zs}}$ & 58.9 & 78.4 & 91.2 \\
$\texttt{Alpaca-7B}_{\texttt{ICL}}$ & 65.2 & 86.9 & 94.8 \\
$\texttt{Vicuna-7B}_{\texttt{zs}}$ & 61.0 & 80.5 & 91.6 \\
$\texttt{Vicuna-7B}_{\texttt{ICL}}$ & 65.8 & 87.4 & 94.3 \\
$\texttt{Falcon-7B}_{\texttt{zs}}$ & 61.9 & 81.0 & 92.1 \\
$\texttt{Falcon-7B}_{\texttt{ICL}}$ & 66.0  & 87.7 & 94.2 \\
\midrule
$\texttt{ChatGPT}_{\texttt{zs}}$ & 66.4 & 84.5 & 97.3\\
$\texttt{ChatGPT}_{\texttt{ICL}}$ & \bf 88.4 & \bf 90.3 & \bf 98.1 \\
\bottomrule
\end{tabular}
\caption{Results on content-constrained text generation.}
\label{tab:content}
\end{table}

\paragraph{Results.}
The results are shown in Table \ref{tab:content}. We find that Alpaca significantly outperforms LLaMA in the zero-shot setting. This is intuitive since natural language instruction of constraints resembles instruction tuning data. However, this performance gap is significantly reduced when in-context learning is used. We think this is because the role of instruction tuning is mainly to adapt an LLM to human-friendly prompt formats instead of increasing the LLM's capability. We also find that \texttt{ChatGPT} achieves competitive performance without in-context learning and outperforms Diffusion-LM, a competitive supervised baseline, by a large margin. Moreover, the performance of \texttt{ChatGPT} can be further improved by adding in-context examples to the prompt. This suggests that LLMs' ability to follow content constraints expressed in natural language depends on three confounding factors: instruction tuning or supervised fine-tuning, in-context learning, and model capacity.

\section{Story Generation}
\label{sec:story_generation}
\begin{table}[t]
\centering
\resizebox{\columnwidth}{!}{%
\begin{tabular}{@{}llccccc@{}}
\toprule
LM & Method & \textbf{rep-2}$\downarrow$ & \textbf{rep-3}$\downarrow$ & \textbf{rep-4}$\downarrow$& \textbf{diversity}$\uparrow$ & \textbf{coherence}$\uparrow$ \\ \midrule
\multicolumn{7}{c}{\bf ROC} \\
 & Human &  1.74&0.32&0.04&0.97&0.48\\\midrule
\parbox[t]{2mm}{\multirow{5}{*}{\rotatebox[origin=c]{90}{GPT-2-XL}}} & Nucleus & 1.80 &0.35  &0.12  &0.97  & 0.33 \\
 & Typical & 2.06 &0.4  &0.16  & 0.97 &0.33  \\
 & $\eta$-sampling & \textbf{0} &  \textbf{0}&  \textbf{0}& \textbf{1.00} &  0.34\\
 & SimCTG &  3.10&0.46  &  0.23& 0.96 &0.32  \\
 & Look-back & 7.24 & 0.92 & 0.14 &0.92  &  0.47\\\midrule
\parbox[t]{2mm}{\multirow{3}{*}{\rotatebox[origin=c]{90}{LLM}}} & Vicuna &2.36  &0.45  &0.15  & 0.97 & 0.60\\
 & Falcon & 2.52 & 1.87 & 1.86 &0.94 & \textbf{0.69} \\
 & \texttt{ChatGPT} & 1.18 &0.10  & 0.02 & 0.98 &  0.52\\\midrule\midrule
\multicolumn{7}{c}{\bf Writing Promts} \\
 & Human &  15.61 &3.78  & 1.24 & 0.80 &0.31  \\\midrule
\parbox[t]{2mm}{\multirow{5}{*}{\rotatebox[origin=c]{90}{GPT-2-XL}}} & Nucleus &5.40 &2.41& 1.72 &0.91&0.34  \\
 & Typical &3.60& 1.51 &1.10 &0.94 &0.30 \\
 & $\eta$-sampling &6.17& 2.88 &2.16& 0.89 &0.35  \\
 & SimCTG & \textbf{2.84} &\textbf{0.36}& \textbf{0.19}& \textbf{0.97} &0.31  \\
 & Look-back & 7.94 &1.25 &0.33& 0.91&0.52  \\\midrule
\parbox[t]{2mm}{\multirow{3}{*}{\rotatebox[origin=c]{90}{LLM}}} & Vicuna &  8.27&  2.59& 1.14 &  0.88&  0.49\\
 & Falcon & 11.20 &  7.79& 6.94 & 0.76&\textbf{0.53}  \\
 & \texttt{ChatGPT} & 5.99& 1.15 & 0.35 &  0.92& 0.52 \\ \bottomrule
\end{tabular}%
}
\caption{Performance of different decoding strategies and LLMs for open-ended story generation. Vicuna stands for Vicuna-7B, Falcon for Falcon-7B-Instruct.
}
\label{tab:open_story_res}
\end{table}

\paragraph{Task Description.}
Given the beginning text of a story, open-ended story generation aims to decode texts that are coherent with previous topics, and informative without undesired repetitions~\citep{su2022contrastive,su2022empirical,xu2023look}. Despite the impressive success on generating fluent and accurate sentences for low-entropy tasks such as summarization or translation, large-scale language models (LLMs) still suffer from serious degeneration problems, such as undesired repetitions~\citep{holtzman2020curious,su2022contrastive} and unnatural topic drifts~\citep{li2022contrastive}, under open-ended settings.

\paragraph{Datasets.}
We evaluate different generation methods on two popular benchmark story datasets: ROCStories and Writing Prompts. ROCStories (ROC)~\citep{mostafazadeh-etal-2016-corpus} is a corpus comprising commonsense stories written by crowd-sourced workers within 5 short sentences.
Given the first sentence as a prefix, generation methods are required to produce four continuing sentences. 
Writing Prompts (WP) is a challenging task for inspiring continuations with abstract, high-level story prompts submitted by online users and continuations by others on Reddit~\citep{fan-etal-2018-hierarchical}. Following prior literature~\citep{xu2023look}, we utilize the first $32$ tokens as the prefix and ask for continuation with $256$ tokens. Since we prompt different language models or decoding algorithms without extra fine-tuning, we directly sample 1,000 development and 1,000 testing instances from both ROC and WP. 


\paragraph{Baselines.}
We evaluate the pre-trained LLM, GPT-2-XL~\citep{radford2019language}, with both search (SimCTG~\citep{su2022contrastive} and Look-back~\citep{xu2023look}) and sampling decoding methods (Nucleus sampling~\citep{holtzman2020curious}, Typical decoding~\citep{meister2022typical} and $\eta$-sampling~\citep{hewitt-etal-2022-truncation}).

\paragraph{Evaluation Metrics.}
Following open-ended story generation literature~\cite{su2022contrastive,li2022contrastive,xu2023look}, we adopt the following automatic metrics to evaluate generation quality: 1) \emph{rep-$n$} to measure sequence-level repetition according to the portion of duplicate n-grams~\citep{welleck2019neural}; 2) \emph{diversity} to assess the overall model repetition by considering \emph{rep-$n$}  at different n-gram levels; 3) \emph{coherence} measured as the cosine similarity between prefix and continuation embeddings represented by SimCSE~\citep{gao-etal-2021-simcse}. 
We do not report MAUVE~\citep{pillutla2021mauve} score due to the concern that MAUVE may not accurately reflect human preferences considering contradicted results between MAUVE and human evaluations observed in prior work~\citep{su2022empirical}.

\paragraph{Evaluate with LLMs.} 
Chatbots that fine-tune LLMs on instructions are also evaluated: Vicuna-7B~\citep{vicuna2023}, Falcon-7B-Instruct~\citep{falcon40b} and \texttt{ChatGPT}.~\footnote{\url{https://chat.openai.com/}} We prepend the following instruction before the story prefix as prompt: 1) ROC: ``Please continue writing this story within 4 very short 
 sentences: <prefix>'', 2) WP: ``Please continue writing this story within 256 words: <prefix>''\footnote{We adopt generation parameters for different LLMs suggested from their respective documents or APIs. We leave evaluation on more configurations in our repository: \url{https://github.com/sunjiao123sun/llm-controlgen}.}.
\paragraph{Results.}
As shown in Table~\ref{tab:open_story_res}, both Vicuna-7B and \texttt{ChatGPT} are able to continue writing more fluent and coherent stories on both ROC and WP compared with other decoding methods based on GPT2-XL. Falcon-7B-Instruct obtains consistently lower diversity than other baselines, while \texttt{ChatGPT} achieves more robust performance in terms of diversity and coherence on both datasets.

\begin{table}[]
\centering
\small
\begin{tabular}{@{}lll@{}}
\toprule
I$\rightarrow$O &\multicolumn{2}{l}{ 0.87} \\
\midrule
I+R$_{\texttt{CoS-E}}\rightarrow$O& \multicolumn{2}{l}{0.92} \\
I+R$_{\texttt{ECQA}}\rightarrow$O& \multicolumn{2}{l}{\textbf{0.99}}\\ \midrule
\textbf{Model} & \textbf{Leakage} & \textbf{Non-Leakage} \\ \midrule
I+R$_{\texttt{Alpaca-7B}}\rightarrow$O  & 0.91 & 0.86 \\
I+R$_{\texttt{LLaMA-7B}}\rightarrow$O  & 0.87 & 0.79 \\ 
I+R$_{\texttt{Vicuna-7B}}\rightarrow$O  & 0.95 & 0.74 \\
I+R$_{\texttt{Falcon-7B}}\rightarrow$O  & 0.83 & 0.65 \\
\midrule
I+R$_{\texttt{ChatGPT}}\rightarrow$O  & \textbf{0.98} & \textbf{0.93} \\
\bottomrule
\end{tabular}
\caption{Rationales generated by \texttt{ChatGPT} are on par with best-crowdsourced rationales ECQA with FlanT5-XXL~\cite{chung2022scaling} as the backbone model. Ruling out leakage results in at least 5\% accuracy drop.}
\label{tab:rationales}
\end{table}

\section{Rationale Generation}
\label{sec:rationale-generation}
\paragraph{Task Description.} Free-form rationales are known to aid model interpretability by providing additional world knowledge or commonsense reasoning steps~\cite{kim2015interactive,lipton2018mythos,AlvarezMelis2018TowardsRI}. ~\citet{cot} show that rationales can improve large language models' ability to solve complex reasoning tasks. Extractive rationales in question-answering tasks are based on the input passage to extract related information to answer the question. Conversely, free-form rationales in the question-answering tasks are open-ended and condition on purely the question and options. \cite{sun-etal-2022-investigating} studies how different the quality of rationales would impact rationales' utilities in terms of improving the model performance and claims that crowdsourced rationales are superior to generated rationales. \citet{sun-etal-2022-investigating} finetunes T5-base for both rationale generation and question answering. With the power of LLMs, we want to revisit the problem and see whether the utility of generated rationales conditioned on the question and options has been improved. 

\paragraph{Evaluation.} We follow previous works and use the performance gap before and after adding rationales in the input to measure the utility of rationales, written as acc(I+R$\rightarrow$O) - acc(I$\rightarrow$O), where I stands for question and options as input, R stands for rationales, and O stands for one of the options as output. For the backbone model for question answering, we use flanT5-XXL~\cite{flan-t5-xxl} instead of T5-base as it can handle longer sequences and is better at reasoning. 

\citet{sun-etal-2022-investigating} shows that two factors are mainly affecting the utility of rationales. One is \emph{leakage}, which means that the correct answer is explicitly written in the rationales, and one can choose the correct answer among all the options by rationales without knowing the questions. The other is \emph{background knowledge}, which is the additional background knowledge or reasoning step that can help answer the question.

\paragraph{Datasets.} CoS-E~\cite{cose} and ECQA~\cite{ecqa} are the most popular free-form rationale datasets through crowdsourcing. ECQA builds on CoS-E and improves the quality of the CoS-E dataset from various aspects, including completeness, comprehensiveness, coherence, etc. They share the same sets of questions and options. Based on the findings from ~\citet{sun-etal-2022-investigating}, both CoS-E and ECQA tend to leak the correct answer in the rationale, while ECQA rationales contain the background necessary to answer the questions. We conduct our analysis on question-answer pairs from the test set. Based on the evaluation acc(I+R$\rightarrow$O) - acc(I$\rightarrow$O), since we are evaluating on the same set of question-answer pairs, acc(I$\rightarrow$O) is always the same. Therefore, we only compare acc(I+R$\rightarrow$O) with different LLMs.

\paragraph{Evaluate with LLMs.} We prompt LLMs to provide background knowledge that can help answer the question and control whether to leak the correct options in rationales. We use \texttt{ChatGPT} as the example for illustration:
\begin{itemize}[leftmargin=*]
    \item \emph{Leakage.} We have \texttt{ChatGPT} take the role of \emph{A teacher who is trying to explain to students the rationale behind choosing the \underline{correct option} for a multiple-choice question.} Then prompt it with \emph{Question: \{question\} Options: \{concatenated options\}  Explain the rationale \underline{behind choosing the correct option} \underline{``\{correct answer\}”}}.
    \item \emph{Non-leakage.} The role of \texttt{ChatGPT} becomes \emph{A teacher who is trying to explain to students the rationale behind a multiple-choice question. \underline{However, you do not want to leak the correct} \underline{answer directly.}} and prompt it with \emph{Question: \{question\} Options: \{concatenated options\} Explain the rationale behind choosing the correct answer. \underline{Do not mention the correct answer} \underline{``\{correct
    answer\}” explicitly}}.
\end{itemize}
We highlight the difference between the two modes with \underline{underline}. When prompting LLaMA and Alpaca, we remove the role description and only use the prompts. Through analysis, we aim to answer two questions: 1) Are LLM-generated rationales on par with crowdsourced rationales? 2) How much would leakage impact the utility of rationales?

\paragraph{Result.} Compared to T5, FlanT5 has better reasoning abilities~\cite{chung2022scaling} and is more capable of understanding instructions. Therefore, we use FlanT5 instead of using T5 as the backbone model for question answering, which can theoretically examine the utility of rationales better ruling out the incapability of models. Simply given the question and the option strings, Table~\ref{tab:rationales} shows that FlanT5-XXL has an accuracy of 0.87 (while T5 in \cite{sun-etal-2022-investigating} scores 0.57 under the same setting). We then show the performance with crowdsourced rationales from both ECQA and CoS-E. With crowdsourced rationales from ECQA, the model almost solved the task and reached a performance of 0.99. With CoS-E rationales, the accuracy is 0.92. Our finding echoes with \citet{sun-etal-2022-investigating} that ECQA rationales are better quality.

We then evaluate the utility of LLM-generated rationales under both the \emph{Leakage} and \emph{Non-leakage} scenarios. As the majority of crowdsourced rationales contain leakage~\cite{sun-etal-2022-investigating}, we consider it fair to compare LLM-generated rationales under the \emph{Leakage} scenarios against crowdsourced rationales. We have two major findings:
\begin{itemize}[leftmargin=*]
    \item \texttt{ChatGPT} generated rationales are on par with ECQA rationales from crowdsourcing.
    \item We quantify the influence of leakage in measuring the utility of rationales: whether or not having leakage in rationales could result in an accuracy difference of at least 5\%. 
\end{itemize} 


\begin{table*}[t]
\small\centering
\begin{tabular}{@{}l|llllllll@{}}
\toprule
& \textbf{} & \textbf{BLEU$\uparrow$} & \textbf{METEOR$\uparrow$} & \textbf{ROUGE-1$\uparrow$} & \textbf{ROUGE-2$\uparrow$} & \textbf{ROUGE-L$\uparrow$} & \textbf{\begin{tabular}[c]{@{}l@{}}TED-R$\downarrow$\\(H=2)\end{tabular}} & \textbf{\begin{tabular}[c]{@{}l@{}}TED-E$\downarrow$\\(H=2)\end{tabular}} \\ \midrule
\multirow{4}{*}{\begin{tabular}[c]{@{}l@{}}ParaNMT\\ -Small\end{tabular}} & Direct & 10.8 & 26.2 & 44.2 & 18.6 & 44.9 & 1.4 & 1.5 \\
 & Ctrl & 14.3 & 30.7 & 51.4 & 25.8 & 50.7 & 1.3 & 1.2 \\
 & Syntax exp. & 13.6 & 27.3 & 46.4 & 20.2 & 47.0 & 1.4 & 1.4 \\ \cmidrule(l){2-9} 
 & {\small\faTrophy} \textbf{AESOP} & \textbf{22.9} & \textbf{32.7} & \textbf{54.4} & \textbf{29.8} & \textbf{56.4} & \textbf{0.9} & \textbf{0.5} \\ \midrule
QQPPos & Direct & 6.7 & 25.2 & 39.8 & 15.6 & 41.5 & 1.8 & 1.8 \\
 & Ctrl & 10.5 & 25.6 & 43.0 & 19.8 & 45.2 & 1.4 & 1.4 \\
 & Syntax exp. & 9.0 & 26.5 & 42.8 & 17.8 & 14.2 & 1.8 & 1.8 \\ \cmidrule(l){2-9} 
 & {\small\faTrophy} \textbf{AESOP} & \textbf{47.3} & \textbf{49.7} & \textbf{73.3} & \textbf{54.1} & \textbf{75.6} & \textbf{0.4} & \textbf{0.3} \\ \bottomrule
\end{tabular}
\caption{Performance comparison with ground-truth syntactic control for AESOP~\cite{sun-etal-2021-aesop} and fine-shot \texttt{ChatGPT}. With coarse syntactic control from a shallow height of pruning, AESOP, the state of the finetuned small model, outperforms five-shot \texttt{ChatGPT} across \textbf{all} semantic preservation (BLUE, ROUGE Scores, and METEOR) and syntactic conformation metrics (TED-R and TED-E at the height of two) by a large margin. $\uparrow$ means higher is better, while $\downarrow$ means lower is better. By comparing \emph{ctrl} with \emph{syntax explanation}, we show that \texttt{ChatGPT} is better at mimicking the syntactic structure from an exemplar than utilizing the syntactic information directly from the syntax. }
\label{tab:paraphrase_performance}
\end{table*}

\section{Controlled Paraphrase Generation}
\label{sec:paraphrase_generation}
\paragraph{Task Description.} Syntactically-controlled paraphrase generation can benefit a wide range of NLP applications such as dialogue generation~\cite{Gao2020ParaphraseAT}, improving the robustness of models~\cite{kuanghao-para} or metrics~\cite{bias-para}, and diversifying other generation tasks such as diverse question generation. 
Syntactically-controlled paraphrase generation is challenging because it requires satisfying two folds of control signals: semantic preservation and syntactic conformation. By definition of paraphrases, the generation should have exactly the same semantics as the input text. With syntax as part of the input, generated paraphrases should also conform with the indicated syntax. The input syntax can come from a variety of sources.

\paragraph{Datasets.} We evaluate on ParaNMT-small~\cite{chena}, derived from ParaNMT~\cite{wieting-gimpel-2018-paranmt}, and QQP-Pos~\cite{tacl}. Our train/dev/test split follows previous works~\cite{tacl, sun-etal-2021-aesop}. Each instance is a tuple of \{source sentence, exemplar, ground-truth paraphrase\}, where the exemplar shares the same syntax with the ground-truth paraphrase. 

\paragraph{Evaluation Metrics.} We use two sets of evaluation metrics to evaluate the quality of generated paraphrases. We use lexical-overlapping-based scores to evaluate the semantic preservation and tree-edit distances to evaluate the syntactic conformation. For lexical-overlapping-based scores, the higher is better. For tree edit distance, the lower is better, indicating that the newly derived syntax matches more closely with the expected syntax. In this work, we prune the constituency parse trees at a level of 2 and only compare the high-level syntactic structure. TED-R means the tree edit distance between the candidate-generated sentence with the ground-truth paraphrase as the reference. TED-E compares the candidate sentence against the exemplar that only provides the syntax. 

\paragraph{Evaluate with LLMs.} We provide three ways to prompt for the controlled paraphrase generation:
\begin{itemize}[leftmargin=*]
    \item \emph{Direct.} We prompt LLMs directly without specifying any constraints. The prompt is written as \emph{Paraphrase \{source sentence\}. Please only have the paraphrase in the response.}
    \item \emph{Control.} Under this mode, we use the exemplar sentence for the syntactic control signal. The prompt is written as \emph{Paraphrase ``\{source sentence\}'' so that it uses the syntactic structure from ``\{exemplar\}''; please only have the paraphrase in the response. }  
\end{itemize}
We observe that under the \emph{Control} mode, the generated paraphrases would sometimes take the syntactic information from the exemplars and the semantic meaning from exemplar sentences. To solve this, we introduce the third mode \emph{Control with syntax explanation}. We first extract the constituency parse structure from the exemplar sentence using Stanford CoreNLP, prune the parse tree at the height of two (i.e., parse at H2), and then ask \texttt{ChatGPT} to generate a natural language explanation of the pruned syntactic parse, which we refer to as \emph{syntax explanation}. The generated syntax explanation will be part of the input. 
\begin{itemize}[leftmargin=*]
    \item  \emph{Control with Syntax Explanation.} The prompt is written as \emph{Paraphrase ``\{source sentence\}" so that the sentence has a syntactic structure of ``\{pruned syntax\}". \{generated explanation for the syntax.\} Please only have the generated paraphrase, not its parse, in the response.}
\end{itemize}

\begin{table}[]
\resizebox{\columnwidth}{!}{
\begin{tabular}{@{}l|l@{}}
\toprule
\textbf{Pruned Parse at H=2} & \textbf{Explanation} \\ \midrule
(ROOT (S (NP ) (VP ))) & \begin{tabular}[c]{@{}l@{}}This represents a sentence structure\\ with a noun phrase and a verb phrase\\  as its constituents.\end{tabular} \\ \midrule
\begin{tabular}[c]{@{}l@{}}(ROOT (FRAG (SBAR )\\  (. )))\end{tabular} & \begin{tabular}[c]{@{}l@{}}This is a sentence with a fragment\\ that includes a subordinate clause \\ followed by a period.\end{tabular} \\ \midrule
\begin{tabular}[c]{@{}l@{}}(ROOT (SBARQ \\ (WHADVP ) (SQ ) (. )))\end{tabular} & \begin{tabular}[c]{@{}l@{}}This sentence structure represents an\\  interrogative sentence with a subord\\ -inate clause before the main clause.\end{tabular} \\ \midrule
\begin{tabular}[c]{@{}l@{}}(ROOT (SQ (VBP ) \\ (RB ) (NP ) (VP ) (. )))\end{tabular} & \begin{tabular}[c]{@{}l@{}}This is a parse tree for a sentence\\ containing a main verb and its subject,\\ with a possible adverb and complement \\ structure.\end{tabular} \\ \bottomrule
\end{tabular}
}
\caption{Examples of generated explanations for pruned constituency parse trees by \texttt{ChatGPT}.}
\label{tab:example_exp}
\end{table}

Table~\ref{tab:example_exp} shows examples of generated explanations for constituency parse trees pruned at height two by \texttt{ChatGPT}. We prompt \texttt{ChatGPT} from zero shots to five shots for our experiments, find that \texttt{ChatGPT}'s performance peaks with five shots as expected, and compare the performance of five-shot \texttt{ChatGPT} with AESOP~\cite{sun-etal-2021-aesop}. The backbone of AESOP is the BART-base model, a 140m-parameter model finetuned with specialized input and output format tailored for the controlled paraphrase generation task. To the best of our knowledge, AESOP remains the state-of-the-art paraphrase generation model on both ParaNMT-small and QQPPos datasets. 

\paragraph{Result.} Table~\ref{tab:paraphrase_performance} shows the performance comparison between five-shot \texttt{ChatGPT} and AESOP. We show that AESOP surpasses \texttt{ChatGPT} across all evaluation metrics for both semantic preservation metrics (lexical-overlapping based metrics including BLEU, ROUGE scores, and METEOR) and syntactic conformation metrics (TED-R and TED-E at the height of two). In addition, we find that \texttt{ChatGPT}'s performance is the best under the setting of \emph{Control}, where we use exemplar sentences for control signals. Compared with the setting \emph{Control with syntax explanation}, Table~\ref{tab:paraphrase_performance} shows that \texttt{ChatGPT} is good at mimicking syntactic structures from sentences instead of directly incorporating the syntactic parses. Besides \texttt{ChatGPT}, we also tried Alpaca~\cite{alpaca} and LLaMA~\cite{llama} on the controlled paraphrase generation task. However, they repeat input sentences and struggle to generate meaningful content. Therefore, we do not include them here for comparison.  

\section{Related Works}
\paragraph{LLM Evaluation.} While the advancement of more potent large language models drives our work, our focus aligns more with recent studies evaluating LLMs' performance on academic NLP benchmarks.
We roughly categorize these studies as either general or specific NLP tasks.  For general NLP tasks, \citet{qin2023chatgpt} shows that \texttt{ChatGPT} performs well on many tasks involving reasoning capabilities but not on sequence tagging. \citet{ahuja2023mega} evaluate LLMs on various multilingual NLP tasks. For specific tasks, ~\citet{jiao2023chatgpt} shows that \texttt{ChatGPT} has achieved competitive performance on machine translation. \citet{gao2023exploring} uses \texttt{ChatGPT} for event extraction and shows that it only matches with around a half percent of specialized event extraction models. To the best of the authors' knowledge, we are the first to study the controllability of LLMs and the tasks in our work have not been previously studied. Instead of having a single conclusion on if LLMs perform well at certain task, we provide a spectrum showcasing how LLMs' abilities vary according to different control granularities.

\section{Discussion: Why and How}
We believe that our work makes a substantial contribution to the field of benchmarking LLMs' controllabiltiy, especially considering the prevalence of LLMs these days. That being said, we do have a few hypotheses to investigate \textbf{\emph{why}} LLMs fail at numerical planning and \textbf{\emph{how}} we could potentially increase their controllability.

\paragraph{Tokenization.} On one hand, tokenization indeed makes the task of numerical planning more challenging than without, by separating the generative process (\textit{i.e.,} subword-level generation) and the numerical planning process (\textit{i.e.,} counting complete words). 
However, we posit that tokenizers not necessarily impact the ability of word planning, as it is a standard practice that a subword starting with a special token will indicate the start of a new word (\textit{e.g.}, “Ġ” in BPE tokenizer,\footnote{\url{https://huggingface.co/learn/nlp-course/chapter6/5?fw=pt##byte-pair-encoding-tokenization}} which has been used by many LLMs such as GPT and RoBERTa). 
Nor are we aware of evidence that the subwords of a tokenizer roughly correspond to units of syllables. For example, \citet{tian-etal-2023-unsupervised} shows that smaller models such as GPT-2-large fine-tuned on syllable-related data can achieve a success rate of close to 90\% on the same syllable-planning task. On the other hand, the best performance of \texttt{ChatGPT} is 37\%.

\paragraph{Decoding Methods.} The reported results are based on sampling with a temperature of 0.3. Moreover, we have experiments showing that our conclusion is robust to the change of decoding mechanisms, where we try other decoding methods beyond sampling with $T=0.3$.

Specifically, we tried 1) greedy decoding, 2) beam search with beam size 8, and 3) sampling with temperature $T=\{0.3, 0.7, 1.0\}$. For the prior two, most of the generated outputs are highly similar, plain, and lack diversity. As for sampling with $T=\{0.3, 0.7, 1.0\}$, the success rate decreases as $T$ increases. We think $T=0.3$ is a reasonable balance between diversity and quality. We believe that our results convey meaningful signals since each number $N$ has been averaged over 100 different evaluation samples to reduce noise. However, none of these experiments show that LLMs can do better than fine-tuned GPT-2.

\paragraph{In-Context Learning.} We try to give more demonstration of NPB in our prompts and we surprisingly found that this does not help once the input number $N$ cannot be found in the examples. Our results resonate with \citet{yin2023did,kung2023models} that LLMs do not truly understand task definitions via in-context learning.

\paragraph{How to Improve.} We encourage future work to explore from two different directions: 1) chain/tree/graph-of-thought reasoning, and 2) bridging LLMs with non-autoregressive generation abilities (e.g., NADO \citep{NEURIPS2022_b40d5797}). For the first one, one can try both simple chain/tree/graph-of-thought prompting or even pretrained LLMs with chain-of-thought/scratchpad pairs, as it shows promises for mathematical reasoning~\citep{zhou2022teaching}. However, this will not fundamentally solve the planning issue. It is straightforward that auto-regressively generating the next tokens will lead to the problem of models not ``looking back'' and therefore not adhering to the fine-grained control signals. Therefore, we encourage researchers to also investigate multi-step planning and iterative revisions with LLMs, or, more fundamentally, challenge the autoregressive architecture of LLMs.

\section{Conclusion}
We test the controllability of large language models on five tasks and ten benchmarks, including a numerical planning benchmark that is easy for humans while challenging for LLMs. From there, we draw a spectrum by comparing the performance between LLMs and smaller specialized models. LLMs are able to generate human-level rationales and conform with coarse control signals, such as sentiment, topic and keyword incorporation.  However, they struggle at fine-grained hard constraints, such as numerical planning and paraphrase generations. We hope that our work can inspire downstream applications on when to adopt LLMs. For example, we find that LLMs are good at generating rationales, and these automatic rationales could be used to further boost LLMs' performance through chain-of-thought reasoning. 

\section*{Acknowledgement}
The authors thank anonymous reviewers for their constructive feedback and suggestions that helped improve the draft, especially reviewer rXWW. Jiao and Yufei are supported by Amazon fellowships. 

\section*{Limitations}
This work is subject to couple of limitations. First, all of our experiments involved heavy prompt engineering effort. Although we have attempted to choose the best performing prompts, there might be room for better prompts which could influence the reported evaluation metrics. Second, automatic evaluations are imperfect. Last, we have not proposed solutions after identifying tasks where LLMs struggle. We leave this for future work.

\bibliography{anthology,custom}

\begin{thebibliography}{64}
\expandafter\ifx\csname natexlab\endcsname\relax\def\natexlab#1{#1}\fi

\bibitem[{Aggarwal et~al.(2022)Aggarwal, Sun, and Peng}]{bias-para}
Arshiya Aggarwal, Jiao Sun, and Nanyun Peng. 2022.
\newblock \href {https://aclanthology.org/2022.findings-emnlp.445} {Towards robust {NLG} bias evaluation with syntactically-diverse prompts}.
\newblock In \emph{Findings of the Association for Computational Linguistics: EMNLP 2022}, pages 6022--6032, Abu Dhabi, United Arab Emirates. Association for Computational Linguistics.

\bibitem[{Aggarwal et~al.(2021)Aggarwal, Mandowara, Agrawal, Khandelwal, Singla, and Garg}]{ecqa}
Shourya Aggarwal, Divyanshu Mandowara, Vishwajeet Agrawal, Dinesh Khandelwal, Parag Singla, and Dinesh Garg. 2021.
\newblock \href {https://doi.org/10.18653/v1/2021.acl-long.238} {{E}xplanations for {C}ommonsense{QA}: {N}ew {D}ataset and {M}odels}.
\newblock In \emph{Proceedings of the 59th Annual Meeting of the Association for Computational Linguistics and the 11th International Joint Conference on Natural Language Processing (Volume 1: Long Papers)}, pages 3050--3065, Online. Association for Computational Linguistics.

\bibitem[{Ahuja et~al.(2023)Ahuja, Diddee, Hada, Ochieng, Ramesh, Jain, Nambi, Ganu, Segal, Axmed, Bali, and Sitaram}]{ahuja2023mega}
Kabir Ahuja, Harshita Diddee, Rishav Hada, Millicent Ochieng, Krithika Ramesh, Prachi Jain, Akshay Nambi, Tanuja Ganu, Sameer Segal, Maxamed Axmed, Kalika Bali, and Sunayana Sitaram. 2023.
\newblock \href {http://arxiv.org/abs/2303.12528} {Mega: Multilingual evaluation of generative ai}.

\bibitem[{Almazrouei et~al.(2023)Almazrouei, Alobeidli, Alshamsi, Cappelli, Cojocaru, Debbah, Goffinet, Heslow, Launay, Malartic, Noune, Pannier, and Penedo}]{falcon40b}
Ebtesam Almazrouei, Hamza Alobeidli, Abdulaziz Alshamsi, Alessandro Cappelli, Ruxandra Cojocaru, Merouane Debbah, Etienne Goffinet, Daniel Heslow, Julien Launay, Quentin Malartic, Badreddine Noune, Baptiste Pannier, and Guilherme Penedo. 2023.
\newblock {Falcon-40B}: an open large language model with state-of-the-art performance.

\bibitem[{Alvarez-Melis and Jaakkola(2018)}]{AlvarezMelis2018TowardsRI}
David Alvarez-Melis and T.~Jaakkola. 2018.
\newblock \href {https://proceedings.neurips.cc/paper/2018/hash/3e9f0fc9b2f89e043bc6233994dfcf76-Abstract.html} {Towards robust interpretability with self-explaining neural networks}.
\newblock In \emph{NeurIPS}.

\bibitem[{Anderson et~al.(2017)Anderson, Fernando, Johnson, and Gould}]{anderson-etal-2017-guided}
Peter Anderson, Basura Fernando, Mark Johnson, and Stephen Gould. 2017.
\newblock \href {https://doi.org/10.18653/v1/D17-1098} {Guided open vocabulary image captioning with constrained beam search}.
\newblock In \emph{Proceedings of the 2017 Conference on Empirical Methods in Natural Language Processing}, pages 936--945, Copenhagen, Denmark. Association for Computational Linguistics.

\bibitem[{Chen et~al.(2019)Chen, Tang, Wiseman, and Gimpel}]{chena}
Mingda Chen, Qingming Tang, Sam Wiseman, and Kevin Gimpel. 2019.
\newblock A multi-task approach for disentangling syntax and semantics in sentence representations.
\newblock pages 2453--2464, Minneapolis, Minnesota. Association for Computational Linguistics.

\bibitem[{Chiang et~al.(2023)Chiang, Li, Lin, Sheng, Wu, Zhang, Zheng, Zhuang, Zhuang, Gonzalez, Stoica, and Xing}]{vicuna2023}
Wei-Lin Chiang, Zhuohan Li, Zi~Lin, Ying Sheng, Zhanghao Wu, Hao Zhang, Lianmin Zheng, Siyuan Zhuang, Yonghao Zhuang, Joseph~E. Gonzalez, Ion Stoica, and Eric~P. Xing. 2023.
\newblock \href {https://lmsys.org/blog/2023-03-30-vicuna/} {Vicuna: An open-source chatbot impressing gpt-4 with 90\%* chatgpt quality}.

\bibitem[{Chung et~al.(2022{\natexlab{a}})Chung, Hou, Longpre, Zoph, Tay, Fedus, Li, Wang, Dehghani, Brahma, Webson, Gu, Dai, Suzgun, Chen, Chowdhery, Narang, Mishra, Yu, Zhao, Huang, Dai, Yu, Petrov, Chi, Dean, Devlin, Roberts, Zhou, Le, and Wei}]{flan-t5-xxl}
Hyung~Won Chung, Le~Hou, Shayne Longpre, Barret Zoph, Yi~Tay, William Fedus, Eric Li, Xuezhi Wang, Mostafa Dehghani, Siddhartha Brahma, Albert Webson, Shixiang~Shane Gu, Zhuyun Dai, Mirac Suzgun, Xinyun Chen, Aakanksha Chowdhery, Sharan Narang, Gaurav Mishra, Adams Yu, Vincent Zhao, Yanping Huang, Andrew Dai, Hongkun Yu, Slav Petrov, Ed~H. Chi, Jeff Dean, Jacob Devlin, Adam Roberts, Denny Zhou, Quoc~V. Le, and Jason Wei. 2022{\natexlab{a}}.
\newblock \href {https://doi.org/10.48550/ARXIV.2210.11416} {Scaling instruction-finetuned language models}.

\bibitem[{Chung et~al.(2022{\natexlab{b}})Chung, Hou, Longpre, Zoph, Tay, Fedus, Li, Wang, Dehghani, Brahma et~al.}]{chung2022scaling}
Hyung~Won Chung, Le~Hou, Shayne Longpre, Barret Zoph, Yi~Tay, William Fedus, Eric Li, Xuezhi Wang, Mostafa Dehghani, Siddhartha Brahma, et~al. 2022{\natexlab{b}}.
\newblock Scaling instruction-finetuned language models.
\newblock \emph{arXiv preprint arXiv:2210.11416}.

\bibitem[{Devlin et~al.(2019)Devlin, Chang, Lee, and Toutanova}]{devlin-etal-2019-bert}
Jacob Devlin, Ming-Wei Chang, Kenton Lee, and Kristina Toutanova. 2019.
\newblock \href {https://doi.org/10.18653/v1/N19-1423} {{BERT}: Pre-training of deep bidirectional transformers for language understanding}.
\newblock In \emph{Proceedings of the 2019 Conference of the North {A}merican Chapter of the Association for Computational Linguistics: Human Language Technologies, Volume 1 (Long and Short Papers)}, pages 4171--4186, Minneapolis, Minnesota. Association for Computational Linguistics.

\bibitem[{Fan et~al.(2018)Fan, Lewis, and Dauphin}]{fan-etal-2018-hierarchical}
Angela Fan, Mike Lewis, and Yann Dauphin. 2018.
\newblock \href {https://doi.org/10.18653/v1/P18-1082} {Hierarchical neural story generation}.
\newblock In \emph{Proceedings of the 56th Annual Meeting of the Association for Computational Linguistics (Volume 1: Long Papers)}, pages 889--898, Melbourne, Australia. Association for Computational Linguistics.

\bibitem[{Gao et~al.(2023)Gao, Zhao, Yu, and Xu}]{gao2023exploring}
Jun Gao, Huan Zhao, Changlong Yu, and Ruifeng Xu. 2023.
\newblock \href {http://arxiv.org/abs/2303.03836} {Exploring the feasibility of chatgpt for event extraction}.

\bibitem[{Gao et~al.(2020)Gao, Zhang, Ou, and Yu}]{Gao2020ParaphraseAT}
Silin Gao, Yichi Zhang, Zhijian Ou, and Zhou Yu. 2020.
\newblock Paraphrase augmented task-oriented dialog generation.
\newblock \emph{ArXiv}, abs/2004.07462.

\bibitem[{Gao et~al.(2021)Gao, Yao, and Chen}]{gao-etal-2021-simcse}
Tianyu Gao, Xingcheng Yao, and Danqi Chen. 2021.
\newblock \href {https://doi.org/10.18653/v1/2021.emnlp-main.552} {{S}im{CSE}: Simple contrastive learning of sentence embeddings}.
\newblock In \emph{Proceedings of the 2021 Conference on Empirical Methods in Natural Language Processing}, pages 6894--6910, Online and Punta Cana, Dominican Republic. Association for Computational Linguistics.

\bibitem[{Hewitt et~al.(2022)Hewitt, Manning, and Liang}]{hewitt-etal-2022-truncation}
John Hewitt, Christopher Manning, and Percy Liang. 2022.
\newblock \href {https://aclanthology.org/2022.findings-emnlp.249} {Truncation sampling as language model desmoothing}.
\newblock In \emph{Findings of the Association for Computational Linguistics: EMNLP 2022}, pages 3414--3427, Abu Dhabi, United Arab Emirates. Association for Computational Linguistics.

\bibitem[{Holtzman et~al.(2020)Holtzman, Buys, Du, Forbes, and Choi}]{holtzman2020curious}
Ari Holtzman, Jan Buys, Li~Du, Maxwell Forbes, and Yejin Choi. 2020.
\newblock \href {http://arxiv.org/abs/1904.09751} {The curious case of neural text degeneration}.

\bibitem[{Huang and Chang(2021)}]{kuanghao-para}
Kuan-Hao Huang and Kai-Wei Chang. 2021.
\newblock Generating syntactically controlled paraphrases without using annotated parallel pairs.
\newblock \emph{ArXiv}, abs/2101.10579.

\bibitem[{Iyyer et~al.(2018)Iyyer, Wieting, Gimpel, and Zettlemoyer}]{iyyer-etal-2018-adversarial}
Mohit Iyyer, John Wieting, Kevin Gimpel, and Luke Zettlemoyer. 2018.
\newblock \href {https://doi.org/10.18653/v1/N18-1170} {Adversarial example generation with syntactically controlled paraphrase networks}.
\newblock In \emph{Proceedings of the 2018 Conference of the North {A}merican Chapter of the Association for Computational Linguistics: Human Language Technologies, Volume 1 (Long Papers)}, pages 1875--1885, New Orleans, Louisiana. Association for Computational Linguistics.

\bibitem[{Jiao et~al.(2023{\natexlab{a}})Jiao, Wang, tse Huang, Wang, and Tu}]{chatgptformt}
Wenxiang Jiao, Wenxuan Wang, Jen tse Huang, Xing Wang, and Zhaopeng Tu. 2023{\natexlab{a}}.
\newblock \href {http://arxiv.org/abs/2301.08745} {Is chatgpt a good translator? yes with gpt-4 as the engine}.

\bibitem[{Jiao et~al.(2023{\natexlab{b}})Jiao, Wang, tse Huang, Wang, and Tu}]{jiao2023chatgpt}
Wenxiang Jiao, Wenxuan Wang, Jen tse Huang, Xing Wang, and Zhaopeng Tu. 2023{\natexlab{b}}.
\newblock \href {http://arxiv.org/abs/2301.08745} {Is chatgpt a good translator? yes with gpt-4 as the engine}.

\bibitem[{Keung et~al.(2020)Keung, Lu, Szarvas, and Smith}]{keung-etal-2020-multilingual}
Phillip Keung, Yichao Lu, Gy{\"o}rgy Szarvas, and Noah~A. Smith. 2020.
\newblock \href {https://doi.org/10.18653/v1/2020.emnlp-main.369} {The multilingual {A}mazon reviews corpus}.
\newblock In \emph{Proceedings of the 2020 Conference on Empirical Methods in Natural Language Processing (EMNLP)}, pages 4563--4568, Online. Association for Computational Linguistics.

\bibitem[{Kim(2015)}]{kim2015interactive}
Been Kim. 2015.
\newblock \href {https://dspace.mit.edu/handle/1721.1/98680} {\emph{Interactive and interpretable machine learning models for human machine collaboration}}.
\newblock Ph.D. thesis, Massachusetts Institute of Technology.

\bibitem[{Kumar et~al.(2020)Kumar, Ahuja, Vadapalli, and Talukdar}]{tacl}
A.~Kumar, Kabir Ahuja, Raghuram Vadapalli, and P.~Talukdar. 2020.
\newblock Syntax-guided controlled generation of paraphrases.
\newblock \emph{Transactions of the Association for Computational Linguistics}, 8:330--345.

\bibitem[{Kung and Peng(2023)}]{kung2023models}
Po-Nien Kung and Nanyun Peng. 2023.
\newblock Do models really learn to follow instructions? an empirical study of instruction tuning.
\newblock \emph{ACL 2023}.

\bibitem[{Laskar et~al.(2023)Laskar, Bari, Rahman, Bhuiyan, Joty, and Huang}]{chatgpt-benchmark}
Md~Tahmid~Rahman Laskar, M~Saiful Bari, Mizanur Rahman, Md~Amran~Hossen Bhuiyan, Shafiq~R. Joty, and J.~Huang. 2023.
\newblock A systematic study and comprehensive evaluation of chatgpt on benchmark datasets.

\bibitem[{Lewis et~al.(2019)Lewis, Liu, Goyal, Ghazvininejad, rahman Mohamed, Levy, Stoyanov, and Zettlemoyer}]{Lewis2019BARTDS}
Mike Lewis, Yinhan Liu, Naman Goyal, Marjan Ghazvininejad, Abdel rahman Mohamed, Omer Levy, Veselin Stoyanov, and Luke Zettlemoyer. 2019.
\newblock Bart: Denoising sequence-to-sequence pre-training for natural language generation, translation, and comprehension.
\newblock In \emph{Annual Meeting of the Association for Computational Linguistics}.

\bibitem[{Li et~al.(2022{\natexlab{a}})Li, Holtzman, Fried, Liang, Eisner, Hashimoto, Zettlemoyer, and Lewis}]{li2022contrastive}
Xiang~Lisa Li, Ari Holtzman, Daniel Fried, Percy Liang, Jason Eisner, Tatsunori Hashimoto, Luke Zettlemoyer, and Mike Lewis. 2022{\natexlab{a}}.
\newblock Contrastive decoding: Open-ended text generation as optimization.
\newblock \emph{arXiv preprint arXiv:2210.15097}.

\bibitem[{Li et~al.(2022{\natexlab{b}})Li, Thickstun, Gulrajani, Liang, and Hashimoto}]{li2022diffusionlm}
Xiang~Lisa Li, John Thickstun, Ishaan Gulrajani, Percy Liang, and Tatsunori Hashimoto. 2022{\natexlab{b}}.
\newblock \href {https://openreview.net/forum?id=3s9IrEsjLyk} {Diffusion-{LM} improves controllable text generation}.
\newblock In \emph{Advances in Neural Information Processing Systems}.

\bibitem[{Lin et~al.(2020)Lin, Zhou, Shen, Zhou, Bhagavatula, Choi, and Ren}]{lin-etal-2020-commongen}
Bill~Yuchen Lin, Wangchunshu Zhou, Ming Shen, Pei Zhou, Chandra Bhagavatula, Yejin Choi, and Xiang Ren. 2020.
\newblock \href {https://doi.org/10.18653/v1/2020.findings-emnlp.165} {{C}ommon{G}en: A constrained text generation challenge for generative commonsense reasoning}.
\newblock In \emph{Findings of the Association for Computational Linguistics: EMNLP 2020}, pages 1823--1840, Online. Association for Computational Linguistics.

\bibitem[{Lipton(2018)}]{lipton2018mythos}
Zachary~C Lipton. 2018.
\newblock \href {https://queue.acm.org/detail.cfm?id=3241340} {The mythos of model interpretability: In machine learning, the concept of interpretability is both important and slippery.}
\newblock \emph{Queue}, 16(3):31--57.

\bibitem[{Lu et~al.(2021)Lu, West, Zellers, Le~Bras, Bhagavatula, and Choi}]{lu-etal-2021-neurologic}
Ximing Lu, Peter West, Rowan Zellers, Ronan Le~Bras, Chandra Bhagavatula, and Yejin Choi. 2021.
\newblock \href {https://doi.org/10.18653/v1/2021.naacl-main.339} {{N}euro{L}ogic decoding: (un)supervised neural text generation with predicate logic constraints}.
\newblock In \emph{Proceedings of the 2021 Conference of the North American Chapter of the Association for Computational Linguistics: Human Language Technologies}, pages 4288--4299, Online. Association for Computational Linguistics.

\bibitem[{Meister et~al.(2022)Meister, Pimentel, Wiher, and Cotterell}]{meister2022typical}
Clara Meister, Tiago Pimentel, Gian Wiher, and Ryan Cotterell. 2022.
\newblock Typical decoding for natural language generation.
\newblock \emph{arXiv preprint arXiv:2202.00666}.

\bibitem[{Meng et~al.(2022)Meng, Lu, Peng, and Chang}]{NEURIPS2022_b40d5797}
Tao Meng, Sidi Lu, Nanyun Peng, and Kai-Wei Chang. 2022.
\newblock \href {https://proceedings.neurips.cc/paper_files/paper/2022/file/b40d5797756800c97f3d525c2e4c8357-Paper-Conference.pdf} {Controllable text generation with neurally-decomposed oracle}.
\newblock In \emph{Advances in Neural Information Processing Systems}, volume~35, pages 28125--28139. Curran Associates, Inc.

\bibitem[{Mostafazadeh et~al.(2016)Mostafazadeh, Chambers, He, Parikh, Batra, Vanderwende, Kohli, and Allen}]{mostafazadeh-etal-2016-corpus}
Nasrin Mostafazadeh, Nathanael Chambers, Xiaodong He, Devi Parikh, Dhruv Batra, Lucy Vanderwende, Pushmeet Kohli, and James Allen. 2016.
\newblock \href {https://doi.org/10.18653/v1/N16-1098} {A corpus and cloze evaluation for deeper understanding of commonsense stories}.
\newblock In \emph{Proceedings of the 2016 Conference of the North {A}merican Chapter of the Association for Computational Linguistics: Human Language Technologies}, pages 839--849, San Diego, California. Association for Computational Linguistics.

\bibitem[{Ouyang et~al.(2022)Ouyang, Wu, Jiang, Almeida, Wainwright, Mishkin, Zhang, Agarwal, Slama, Ray, Schulman, Hilton, Kelton, Miller, Simens, Askell, Welinder, Christiano, Leike, and Lowe}]{ouyang2022training}
Long Ouyang, Jeff Wu, Xu~Jiang, Diogo Almeida, Carroll~L. Wainwright, Pamela Mishkin, Chong Zhang, Sandhini Agarwal, Katarina Slama, Alex Ray, John Schulman, Jacob Hilton, Fraser Kelton, Luke Miller, Maddie Simens, Amanda Askell, Peter Welinder, Paul Christiano, Jan Leike, and Ryan Lowe. 2022.
\newblock \href {http://arxiv.org/abs/2203.02155} {Training language models to follow instructions with human feedback}.

\bibitem[{Pillutla et~al.(2021)Pillutla, Swayamdipta, Zellers, Thickstun, Welleck, Choi, and Harchaoui}]{pillutla2021mauve}
Krishna Pillutla, Swabha Swayamdipta, Rowan Zellers, John Thickstun, Sean Welleck, Yejin Choi, and Zaid Harchaoui. 2021.
\newblock Mauve: Measuring the gap between neural text and human text using divergence frontiers.
\newblock \emph{Advances in Neural Information Processing Systems}, 34:4816--4828.

\bibitem[{Post and Vilar(2018)}]{post-vilar-2018-fast}
Matt Post and David Vilar. 2018.
\newblock \href {https://doi.org/10.18653/v1/N18-1119} {Fast lexically constrained decoding with dynamic beam allocation for neural machine translation}.
\newblock In \emph{Proceedings of the 2018 Conference of the North {A}merican Chapter of the Association for Computational Linguistics: Human Language Technologies, Volume 1 (Long Papers)}, pages 1314--1324, New Orleans, Louisiana. Association for Computational Linguistics.

\bibitem[{Qian et~al.(2019)Qian, Qiu, Zhang, Jiang, and Yu}]{qian-etal-2019-exploring}
Lihua Qian, Lin Qiu, Weinan Zhang, Xin Jiang, and Yong Yu. 2019.
\newblock \href {https://doi.org/10.18653/v1/D19-1313} {Exploring diverse expressions for paraphrase generation}.
\newblock In \emph{Proceedings of the 2019 Conference on Empirical Methods in Natural Language Processing and the 9th International Joint Conference on Natural Language Processing (EMNLP-IJCNLP)}, pages 3173--3182, Hong Kong, China. Association for Computational Linguistics.

\bibitem[{Qin et~al.(2023)Qin, Zhang, Zhang, Chen, Yasunaga, and Yang}]{qin2023chatgpt}
Chengwei Qin, Aston Zhang, Zhuosheng Zhang, Jiaao Chen, Michihiro Yasunaga, and Diyi Yang. 2023.
\newblock \href {http://arxiv.org/abs/2302.06476} {Is chatgpt a general-purpose natural language processing task solver?}

\bibitem[{Qin et~al.(2022)Qin, Welleck, Khashabi, and Choi}]{qin2022cold}
Lianhui Qin, Sean Welleck, Daniel Khashabi, and Yejin Choi. 2022.
\newblock \href {https://openreview.net/forum?id=TiZYrQ-mPup} {{COLD} decoding: Energy-based constrained text generation with langevin dynamics}.
\newblock In \emph{Advances in Neural Information Processing Systems}.

\bibitem[{Radford et~al.(2019)Radford, Wu, Child, Luan, Amodei, Sutskever et~al.}]{radford2019language}
Alec Radford, Jeffrey Wu, Rewon Child, David Luan, Dario Amodei, Ilya Sutskever, et~al. 2019.
\newblock Language models are unsupervised multitask learners.
\newblock \emph{OpenAI blog}, 1(8):9.

\bibitem[{Rajani et~al.(2019)Rajani, McCann, Xiong, and Socher}]{cose}
Nazneen~Fatema Rajani, Bryan McCann, Caiming Xiong, and Richard Socher. 2019.
\newblock \href {https://arxiv.org/abs/1906.02361} {Explain yourself! leveraging language models for commonsense reasoning}.
\newblock In \emph{Proceedings of the 2019 Conference of the Association for Computational Linguistics (ACL2019)}.

\bibitem[{Reid et~al.(2022)Reid, Zhong, Gururangan, and Zettlemoyer}]{reid-etal-2022-m2d2}
Machel Reid, Victor Zhong, Suchin Gururangan, and Luke Zettlemoyer. 2022.
\newblock \href {https://aclanthology.org/2022.emnlp-main.63} {{M}2{D}2: A massively multi-domain language modeling dataset}.
\newblock In \emph{Proceedings of the 2022 Conference on Empirical Methods in Natural Language Processing}, pages 964--975, Abu Dhabi, United Arab Emirates. Association for Computational Linguistics.

\bibitem[{Sinha et~al.(2023)Sinha, Gauthier, Mueller, Misra, Fuentes, Levy, and Williams}]{sinha-etal-2023-language}
Koustuv Sinha, Jon Gauthier, Aaron Mueller, Kanishka Misra, Keren Fuentes, Roger Levy, and Adina Williams. 2023.
\newblock \href {https://doi.org/10.18653/v1/2023.acl-long.333} {Language model acceptability judgements are not always robust to context}.
\newblock In \emph{Proceedings of the 61st Annual Meeting of the Association for Computational Linguistics (Volume 1: Long Papers)}, pages 6043--6063, Toronto, Canada. Association for Computational Linguistics.

\bibitem[{Smith et~al.(2020)Smith, Gonzalez-Rico, Dinan, and Boureau}]{smith2020controlling}
Eric~Michael Smith, Diana Gonzalez-Rico, Emily Dinan, and Y-Lan Boureau. 2020.
\newblock \href {http://arxiv.org/abs/2009.10855} {Controlling style in generated dialogue}.

\bibitem[{Su et~al.(2022)Su, Lan, Wang, Yogatama, Kong, and Collier}]{su2022contrastive}
Yixuan Su, Tian Lan, Yan Wang, Dani Yogatama, Lingpeng Kong, and Nigel Collier. 2022.
\newblock A contrastive framework for neural text generation.
\newblock \emph{arXiv preprint arXiv:2202.06417}.

\bibitem[{Su and Xu(2022)}]{su2022empirical}
Yixuan Su and Jialu Xu. 2022.
\newblock An empirical study on contrastive search and contrastive decoding for open-ended text generation.
\newblock \emph{arXiv preprint arXiv:2211.10797}.

\bibitem[{Sun et~al.(2021)Sun, Ma, and Peng}]{sun-etal-2021-aesop}
Jiao Sun, Xuezhe Ma, and Nanyun Peng. 2021.
\newblock \href {https://doi.org/10.18653/v1/2021.emnlp-main.420} {{AESOP}: Paraphrase generation with adaptive syntactic control}.
\newblock In \emph{Proceedings of the 2021 Conference on Empirical Methods in Natural Language Processing}, pages 5176--5189, Online and Punta Cana, Dominican Republic. Association for Computational Linguistics.

\bibitem[{Sun et~al.(2022)Sun, Swayamdipta, May, and Ma}]{sun-etal-2022-investigating}
Jiao Sun, Swabha Swayamdipta, Jonathan May, and Xuezhe Ma. 2022.
\newblock \href {https://aclanthology.org/2022.findings-emnlp.432} {Investigating the benefits of free-form rationales}.
\newblock In \emph{Findings of the Association for Computational Linguistics: EMNLP 2022}, pages 5867--5882, Abu Dhabi, United Arab Emirates. Association for Computational Linguistics.

\bibitem[{Taori et~al.(2023)Taori, Gulrajani, Zhang, Dubois, Li, Guestrin, Liang, and Hashimoto}]{alpaca}
Rohan Taori, Ishaan Gulrajani, Tianyi Zhang, Yann Dubois, Xuechen Li, Carlos Guestrin, Percy Liang, and Tatsunori~B. Hashimoto. 2023.
\newblock Stanford alpaca: An instruction-following llama model.
\newblock \url{https://github.com/tatsu-lab/stanford_alpaca}.

\bibitem[{Tian et~al.(2023)Tian, Narayan-Chen, Oraby, Cervone, Sigurdsson, Tao, Zhao, Chung, Huang, and Peng}]{tian-etal-2023-unsupervised}
Yufei Tian, Anjali Narayan-Chen, Shereen Oraby, Alessandra Cervone, Gunnar Sigurdsson, Chenyang Tao, Wenbo Zhao, Tagyoung Chung, Jing Huang, and Nanyun Peng. 2023.
\newblock \href {https://doi.org/10.18653/v1/2023.acl-long.513} {Unsupervised melody-to-lyrics generation}.
\newblock In \emph{Proceedings of the 61st Annual Meeting of the Association for Computational Linguistics (Volume 1: Long Papers)}, pages 9235--9254, Toronto, Canada. Association for Computational Linguistics.

\bibitem[{Tian and Peng(2022)}]{tian-peng-2022-zero}
Yufei Tian and Nanyun Peng. 2022.
\newblock \href {https://doi.org/10.18653/v1/2022.naacl-main.262} {Zero-shot sonnet generation with discourse-level planning and aesthetics features}.
\newblock In \emph{Proceedings of the 2022 Conference of the North American Chapter of the Association for Computational Linguistics: Human Language Technologies}, pages 3587--3597, Seattle, United States. Association for Computational Linguistics.

\bibitem[{Touvron et~al.(2023)Touvron, Lavril, Izacard, Martinet, Lachaux, Lacroix, Rozi{\`e}re, Goyal, Hambro, Azhar et~al.}]{llama}
Hugo Touvron, Thibaut Lavril, Gautier Izacard, Xavier Martinet, Marie-Anne Lachaux, Timoth{\'e}e Lacroix, Baptiste Rozi{\`e}re, Naman Goyal, Eric Hambro, Faisal Azhar, et~al. 2023.
\newblock Llama: Open and efficient foundation language models.
\newblock \emph{arXiv preprint arXiv:2302.13971}.

\bibitem[{Wei et~al.(2022)Wei, Wang, Schuurmans, Bosma, Chi, Le, and Zhou}]{cot}
Jason Wei, Xuezhi Wang, Dale Schuurmans, Maarten Bosma, Ed~H. Chi, Quoc Le, and Denny Zhou. 2022.
\newblock \href {http://arxiv.org/abs/2201.11903} {Chain of thought prompting elicits reasoning in large language models}.
\newblock \emph{CoRR}, abs/2201.11903.

\bibitem[{Welleck et~al.(2019)Welleck, Kulikov, Roller, Dinan, Cho, and Weston}]{welleck2019neural}
Sean Welleck, Ilia Kulikov, Stephen Roller, Emily Dinan, Kyunghyun Cho, and Jason Weston. 2019.
\newblock Neural text generation with unlikelihood training.
\newblock \emph{arXiv preprint arXiv:1908.04319}.

\bibitem[{Wieting and Gimpel(2018)}]{wieting-gimpel-2018-paranmt}
John Wieting and Kevin Gimpel. 2018.
\newblock \href {https://doi.org/10.18653/v1/P18-1042} {{P}ara{NMT}-50{M}: Pushing the limits of paraphrastic sentence embeddings with millions of machine translations}.
\newblock In \emph{Proceedings of the 56th Annual Meeting of the Association for Computational Linguistics (Volume 1: Long Papers)}, pages 451--462, Melbourne, Australia. Association for Computational Linguistics.

\bibitem[{Xu et~al.(2023{\natexlab{a}})Xu, Zhou, Celikyilmaz, and Ma}]{xu2023lookback}
Nan Xu, Chunting Zhou, Asli Celikyilmaz, and Xuezhe Ma. 2023{\natexlab{a}}.
\newblock \href {http://arxiv.org/abs/2305.13477} {Look-back decoding for open-ended text generation}.

\bibitem[{Xu et~al.(2023{\natexlab{b}})Xu, Zhou, Celikyilmaz, and Ma}]{xu2023look}
Nan Xu, Chunting Zhou, Asli Celikyilmaz, and Xuezhe Ma. 2023{\natexlab{b}}.
\newblock Look-back decoding for open-ended text generation.
\newblock \emph{arXiv preprint arXiv:2305.13477}.

\bibitem[{Yin et~al.(2023)Yin, Vig, Laban, Joty, Xiong, and Wu}]{yin2023did}
Fan Yin, Jesse Vig, Philippe Laban, Shafiq Joty, Caiming Xiong, and Chien-Sheng~Jason Wu. 2023.
\newblock Did you read the instructions? rethinking the effectiveness of task definitions in instruction learning.
\newblock \emph{ACL 2023}.

\bibitem[{Zhang et~al.(2022)Zhang, Song, Li, Zhou, and Song}]{Zhang2022ASO}
Hanqing Zhang, Haolin Song, Shaoyu Li, Ming Zhou, and Dawei Song. 2022.
\newblock A survey of controllable text generation using transformer-based pre-trained language models.
\newblock \emph{ArXiv}, abs/2201.05337.

\bibitem[{Zhou et~al.(2022)Zhou, Nova, Larochelle, Courville, Neyshabur, and Sedghi}]{zhou2022teaching}
Hattie Zhou, Azade Nova, Hugo Larochelle, Aaron Courville, Behnam Neyshabur, and Hanie Sedghi. 2022.
\newblock \href {http://arxiv.org/abs/2211.09066} {Teaching algorithmic reasoning via in-context learning}.

\bibitem[{Zhou et~al.(2023)Zhou, Jiang, Wilcox, Cotterell, and Sachan}]{zhou2023controlled}
Wangchunshu Zhou, Yuchen~Eleanor Jiang, Ethan Wilcox, Ryan Cotterell, and Mrinmaya Sachan. 2023.
\newblock \href {http://arxiv.org/abs/2304.14293} {Controlled text generation with natural language instructions}.

\bibitem[{Çağlayan and Karakaya(2021)}]{9558910}
Cansen Çağlayan and Murat Karakaya. 2021.
\newblock \href {https://doi.org/10.1109/UBMK52708.2021.9558910} {Topic-controlled text generation}.
\newblock In \emph{2021 6th International Conference on Computer Science and Engineering (UBMK)}, pages 533--536.

\end{thebibliography}
\bibliographystyle{acl_natbib}

\newpage
\appendix
\section{SPB additional results}\label{appendix:addtional_SPB}

\begin{table}[t]
\small\centering
\begin{tabular}{@{}lllll@{}}
\toprule
\textbf{Model} & \textbf{\begin{tabular}[c]{@{}l@{}}SR -\\ count\end{tabular}} & \textbf{\begin{tabular}[c]{@{}l@{}}SR -\\ suffix\end{tabular}} & \textbf{\begin{tabular}[c]{@{}l@{}}SR -\\ both\end{tabular}} & \textbf{\begin{tabular}[c]{@{}l@{}}MSE -\\ count\end{tabular}} \\ \midrule
\multicolumn{5}{c}{\cellcolor[HTML]{EFEFEF}syllable planning}  \\
\texttt{ChatGPT}       & 0.37                 & 0.75                     & 0.32                & 4.83        \\
\texttt{ChatGPT}   ICL & 0.30                 & 0.84                     & 0.28                & 6.10        \\
Alpaca-7b     & 0.15                 & 0.33                     & 0.07                & 9.44        \\
Alpaca-7b ICL & 0.12                 & 0.36                     & 0.05                & 10.61       \\ \midrule
\multicolumn{5}{c}{\cellcolor[HTML]{EFEFEF}sentence planning}  \\
\texttt{ChatGPT}       & 0.38 & 0.625 & 0.29 & 1.69 \\
\texttt{ChatGPT}   ICL & 0.36 & 0.66  & 0.27 & 2.05 \\
Alpaca-7b     & 0.19 & 0.19  & 0.07 & 6.56 \\
Alpaca-7b ICL & 0.17 & 0.26  & 0.10 & 8.04 \\ \midrule
\multicolumn{5}{c}{\cellcolor[HTML]{EFEFEF}paragraph planning}  \\
\texttt{ChatGPT}       & 0.69                 & 0.17                  & 0.                  & 3.24        \\
\texttt{ChatGPT}   ICL & 0.57                 & 0.17                  & 0.34                & 4.43        \\
Alpaca-7b     & Failed               &                       &                     &             \\
Alpaca-7b ICL & Failed               &                       &                     &             \\ \bottomrule

\end{tabular}
\caption{Success rates for the syllable, sentence, and paragraph count planning tasks. LLMs are best at sentence count planning and worst at syllable count planning.}
\label{tab:additional_count_planning}
\end{table}

We report the additional results of \texttt{ChatGPT} and Alpaca on the SPB benchmark in Table \ref{tab:additional_count_planning}. Recall that the suffix for the paragraph planning task is the last sentence. In practice, LLMs are unable to follow instructions and copy the requirement as prompted. Hence, when we compute the success rate for this last task, we check the token overlap between the generated sentence and our requirement, and if more than 2/3 of the tokens overlap, we will consider it as a success.

Taking all four tasks in the SPB benchmark into account, we find out that Alpaca-7b have very little numerical planning ability. \texttt{ChatGPT} on the hother hand is best at sentence count planning, and worst at syllable count planning.
\section{Additional Information of Content Controlled Generation}\label{appendix:addtional_content_control}
Controlled content generation refers to the task of controlling the content of generated texts. We consider three types of content constraints:
\begin{itemize}[leftmargin=*]
\vspace{-0.05in}
\itemsep-.5em
    \item \textit{Topic constraint.} It requires the model to generate texts about certain topics. Traditional methods for topic constrained generation either append a special token for different topics~\citep{9558910} or use trained topic classifiers~\citep{qin2022cold} to guide the generation process. 
    \item \textit{Sentiment constraint.} Similar to topic constraint, this task requires the model to generate texts of certain sentiments. The aforementioned methods for topic constrained generation also apply to sentiment constrained generation.
    \item \textit{Keyword constraint.} Keyword constrained, or lexical constrained text generation requires the model to generate texts that contain certain keywords or tokens. Traditional methods for keyword constrained text generation generally enforce lexical constraints on the outputs by modifying the search space according to the constraints~\citep{anderson-etal-2017-guided,post-vilar-2018-fast,lu-etal-2021-neurologic}.
\end{itemize}

\paragraph{Datasets.}
For topic constraints, we use a subset of the topics from the first hierarchy in the M2D2 dataset~\citep{reid-etal-2022-m2d2} which contains domains such as health, history, society, technology, arts, science, etc. The total number of topics is 10 in our experiments. We use the Amazon Review dataset~\citep{keung-etal-2020-multilingual} for sentiment constrained text generation. The sentiment is measure by 1 to 5 stars. For lexical constrained text generation, we use the CommonGEN dataset~\citep{lin-etal-2020-commongen} which requires the model to generate a sentence using three to five keywords.




\end{document}